%% file: main.tex
\makeatletter
\def\input@path{{template/}{./template/}{tex/}{./tex/}}
\makeatother

\documentclass[runningheads]{llncs}

\usepackage{accv}

\usepackage{accvabbrv}

\usepackage{graphicx}
\usepackage{booktabs}
\usepackage{colortbl}  
\usepackage{amsmath}
\usepackage{amssymb}

\usepackage[accsupp]{axessibility}  

\input{tex/macros.tex}

\graphicspath{{figures/}{figures/qualitative/}{figures/analysis/}}

\usepackage[breaklinks,colorlinks,citecolor=accvblue]{hyperref}

\usepackage{orcidlink}
\usepackage{pifont}   

\begin{document}

\title{PoE-Fuse: Precision-Weighted Expert Fusion\\ for Bi-Temporal Change Understanding}

\titlerunning{PoE-Fuse for Bi-Temporal EO Change Understanding}

\author{Haruki Watase\thanks{Equal contribution.}\orcidlink{0009-0008-4029-5735}\textsuperscript{(\ding{41})} \and Shunya Nagashima\protect\footnotemark[1]\orcidlink{0009-0007-9741-7628} \and Takayuki Nishimura\orcidlink{0009-0006-5261-5754}}

\authorrunning{H.~Watase et al.}

\institute{Neurogica Inc., Tokyo, Japan\\
\email{\{haruki.watase,shunya.nagashima,takayuki.nishimura\}@neurogica.com}}

\maketitle

\begin{abstract}
  \looseness=-1
  Bi-temporal change understanding, which localizes and characterizes what changed between two satellite images, is central to disaster response and environmental monitoring, spanning change detection, building localization, and damage assessment.
  Strong vision-language models address these tasks, but adapting them typically requires full fine-tuning or reinforcement learning, which is costly and unstable.
  We propose \method{}, a parameter-efficient framework that instead composes frozen foundation experts for geometry, grounding, and language, resampling their features onto a shared spatial grid and training only a lightweight fusion trunk.
  \method{} treats the aligned features as Gaussian observations of a latent scene state and fuses them by learned per-cell precision.
  This product-of-experts estimator strictly generalizes uniform summation and scalar gating, and extends to change fields by composing the precisions of the two timestamps.
  A single shared trunk solves the three tasks at once, reaching a mean F1 of 59.2\%, compared with 40.7\% for an instruction-tuned temporal vision-language assistant, and surpassing dedicated change-detection models \rev{retrained under the same protocol and training budget}.
\end{abstract}

\input{tex/sections/01_introduction}
\input{tex/sections/02_related_work}
\input{tex/sections/03_problemsettings}
\input{tex/sections/04_method}
\input{tex/sections/05_experiments}
\input{tex/sections/06_conclusion}


%
%
\bibliographystyle{template/splncs04}
\bibliography{main}

%
\clearpage
\section*{Supplementary Material}
\input{tex/sections/supp_body}
\end{document}

%% file: tex/macros.tex
\newcommand{\method}{PoE-Fuse}

\colorlet{revcolor}{black}
\DeclareRobustCommand{\rev}[1]{{\color{revcolor}#1}}



%% file: tex/sections/01_introduction.tex
\section{Introduction}
\label{sec:intro}

\begin{figure}[tb]
  \centering
  \includegraphics[width=\linewidth]{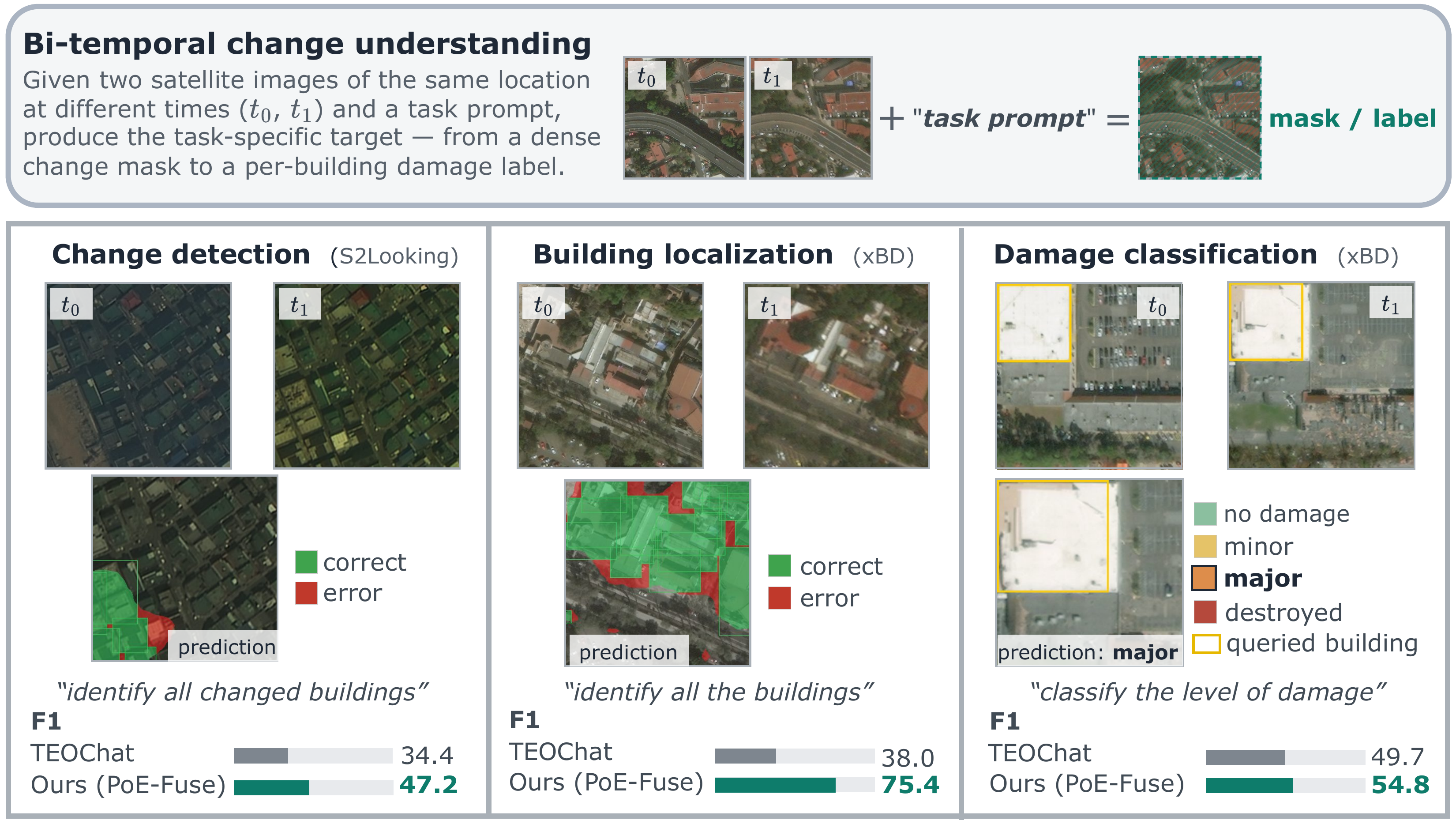}
  \caption{Bi-temporal change understanding and \method{}'s results.
    Given two co-registered satellite images and a task prompt, the model produces a task-specific target: a change mask (S2Looking change detection), building footprints (xBD localization), or a per-building damage label (xBD damage classification).
    Green/red mark correct/wrong predictions; bars compare \method{} with TEOChat~\cite{irvin2025teochat} per task.
  }
  \label{fig:teaser}
\end{figure}

Remote sensing observes the Earth's surface over wide areas at regular revisit intervals, and comparing co-registered satellite images acquired at different times supports disaster response, urban monitoring, and environmental assessment by localizing where the surface changed and, when needed, characterizing it under language guidance.
For instance, in the 2023 T\"urkiye earthquakes, building collapse caused most of the over 41{,}000 deaths~\cite{worldbank2023turkiye}, and the ``golden 72-hour'' window for search and rescue~\cite{chiu2020sar} leaves little time to assess damage across zones beyond ground or drone inspection, making automatic satellite-based assessment essential.

In \emph{bi-temporal change understanding} (Fig.~\ref{fig:teaser}), a model maps two co-registered images of the same location and a short task prompt to a task-specific target such as a change mask or a building damage class.
This setting remains difficult because genuine changes are often small and local, while illumination, season, and residual misregistration create \emph{spurious change} that is easy to confuse with semantic change, and the categories of interest differ from one task to another.

Existing methods cover this setting only partially.
Instruction-tuned temporal assistants such as TEOChat~\cite{irvin2025teochat} adapt large vision-language models (VLMs) to Earth observation (EO) sequences through task-specific fine-tuning.
Single-image EO chat models such as GeoChat~\cite{kuckreja2024geochat} ground language in one timestamp but are not designed to reason across time.
Reasoning-oriented pipelines add supervised fine-tuning and reinforcement learning with process rewards~\cite{yao2026remotereasoner,fiaz2025geovlmr1,sun2026geosolver,shu2026terrascope}, which can be costly and unstable to train.
Geospatial foundation encoders~\cite{jakubik2023foundation,klemmer2025satclip} provide strong representations yet still rely on downstream adaptation for dense change outputs.
Classical change detectors~\cite{chen2022bit,bandara2022changeformer} focus on pixels but do not unify heterogeneous multimodal experts under one geometry.

To address this gap, we propose \method{}\footnote{\rev{Code is available at \url{https://github.com/Neurogica/PoE-Fuse}.}}, which fuses frozen heterogeneous foundation experts.
Unlike approaches that gain generality by fine-tuning a single model or adapting it with reinforcement learning, \method{} keeps every expert frozen, resamples each onto the spatial frame of a geometric anchor, fuses the aligned features and their bi-temporal differences by learned per-cell precision, and passes the result to a lightweight mixer and per-task heads.

Two ideas drive this design.
First, the experts differ in reliability from cell to cell (one is confident on building edges, another on homogeneous texture), so weighting their per-cell evidence by a learned precision combines it more faithfully than a fixed or uniform rule.
Second, resampling the experts onto a common geometric grid is what makes this per-cell fusion well defined, since a grid coordinate then refers to the same location across experts and timestamps.

Our main contributions are as follows:
\begin{itemize}
  \item We propose \method{}, a precision-weighted fusion of frozen heterogeneous experts that treats their aligned features as Gaussian observations of a shared latent scene state and combines them by learned per-cell precision, strictly generalizing uniform summation and scalar gating, with a difference-precision corollary for change fields.
  \item We show that a single shared trunk built on this fusion solves three tasks at once, reaching a three-task mean F1 of $59.2\%$ against $40.7\%$ for an instruction-tuned temporal VLM and surpassing dedicated change-detection models \rev{retrained under the same protocol and training budget}.
  \item Ablations over fusion rules, coupling sites, and expert subsets show that dynamic per-cell precision \rev{improves over its uniform and static special cases, with scalar gating within seed noise,} and that the geometric anchor is indispensable for dense change prediction.
\end{itemize}

%% file: tex/sections/02_related_work.tex
\section{Related Work}
\label{sec:related}

\subsection{Vision-Language Models for Earth Observation}

GeoChat~\cite{kuckreja2024geochat} and related single-image chat models ground natural-language instructions in one satellite snapshot, often building on domain-aligned contrastive pretraining such as RemoteCLIP~\cite{liu2024remoteclip}.
TEOChat~\cite{irvin2025teochat} extends this line to temporal EO sequences by instruction-tuning a video-style assistant on bi-temporal image pairs and task prompts, relying on task-specific adaptation of a multi-billion-parameter backbone.
Parallel geospatial foundation models, surveyed by Jakubik et al.~\cite{jakubik2023foundation} and instantiated in location-aware encoders such as SatCLIP~\cite{klemmer2025satclip}, offer scalable self-supervised representations for downstream mapping and classification.
These approaches typically center on one dominant encoder and substantial fine-tuning or task-specific heads, which makes it difficult to compose geometry, detection, and language backbones for dense bi-temporal outputs without repeatedly updating large weights.
Pixel-level change detection forms a parallel lineage of dedicated siamese and transformer networks~\cite{chen2022bit,bandara2022changeformer,shi2020cdsurvey}, with recent vision-language extensions that tie change maps to text~\cite{li2026btcchat,qu2026pm3net,liu2024rscama}; these train a single change-specific model rather than fusing heterogeneous frozen experts.

Recent reasoning-oriented systems couple supervised fine-tuning with reinforcement learning (RL) and process-level rewards for Earth observation~\cite{yao2026remotereasoner,fiaz2025geovlmr1,sun2026geosolver,shu2026terrascope}.
These pipelines multiply training stages, depend on curated reward signals, and are hard to reproduce at modest compute budgets.
\method{} instead trains only a lightweight fusion stack on cached features of frozen experts, without RL or full-model fine-tuning, and serves several change tasks with one shared trunk rather than one fine-tuned model per task.

\subsection{Frozen-Encoder Fusion and Spatial Alignment}

Freezing large pretrained encoders while training a small interface has become standard in multimodal learning.
BLIP-2~\cite{li2023blip2} bridges a frozen image encoder to a frozen language model with a query transformer, Flamingo~\cite{alayrac2022flamingo} inserts gated cross-attention adapters between frozen vision and language towers, and Frozen CLIP~\cite{lin2022frozenclip} shows that a fixed CLIP backbone can support efficient video understanding with lightweight temporal modules.
Perceiver~\cite{jaegle2021perceiver} and Perceiver IO~\cite{jaegle2022perceiverio} resample variable-length inputs into a fixed latent bottleneck through cross-attention, enabling fusion without modifying upstream weights.
Segment Anything~\cite{kirillov2023sam}, SAM~2~\cite{ravi2025sam2}, and language-conditioned variants such as LISA~\cite{lai2024lisa} provide strong spatial priors from frozen or lightly adapted segmentation backbones.
Across these recipes, resampling is usually one-way from a single vision source into a language latent or a pooled token set, and the output emphasis is often sentence-level text or sparse prompts rather than a dense change field over a shared grid.
When several encoders are combined, their features are typically merged by concatenation, addition, or a learned scalar gate, which weight the experts uniformly or with a single coefficient per expert.
Product-of-experts combinations of probabilistic experts date back to Hinton~\cite{hinton2002poe}, Gaussian products underpin multimodal generative models~\cite{wu2018mvae}, and learned inverse-variance weighting is standard for heteroscedastic uncertainty~\cite{kendall2017uncertainty}; we adapt this lineage to per-cell fusion of frozen foundation experts for dense bi-temporal prediction.

\method{} differs in how the experts are fused: it treats each expert's per-cell feature as a noisy observation of a shared latent scene and weights it by a learned per-cell precision (inverse variance), the best linear unbiased combination, which subsumes uniform averaging and scalar gating; resampling onto the frame of a geometric anchor makes this per-cell fusion well defined.
Among the closest prior methods, TEOChat~\cite{irvin2025teochat} remains our primary benchmark and adapts a 7B-scale temporal assistant through task-specific fine-tuning, whereas we train only a lightweight fusion stack on cached features without updating the experts.
TerraScope~\cite{shu2026terrascope} pursues pixel-grounded EO reasoning yet still depends on large-scale supervised adaptation and curated training data beyond lightweight fusion on cached penultimate features.

%% file: tex/sections/03_problemsettings.tex
\section{Problem Setting}
\label{sec:problem}

We address bi-temporal change understanding for Earth observation on three standard tasks of the TEOChat benchmark~\cite{irvin2025teochat}, namely change detection on S2Looking~\cite{shen2021s2looking}, and building localization and building damage classification on xBD~\cite{gupta2019xbd}, with a change mask, a building-localization mask, and a damage label as the respective outputs.
The input is a co-registered image pair $x_{t_0}, x_{t_1} \in \mathbb{R}^{3 \times H \times W}$ with $H = W = 224$, acquired at times $t_0 < t_1$ over the same location, together with a task prompt: a short instruction for the two mask tasks and a question for damage classification.
Here, co-registered means that the two images share a pixel grid, so a given coordinate refers to the same ground location at $t_0$ and $t_1$, and spurious change denotes an apparent change caused by illumination, seasonal, or registration differences rather than a genuine change of the land surface.
We assume a co-registered pair ($T = 2$) at this fixed resolution; longer acquisition sequences and unregistered inputs are out of scope.
We use pixel-level F1 for change detection and localization and inverse-prevalence-weighted F1 for damage classification, following the benchmark protocol~\cite{irvin2025teochat}.

%% file: tex/sections/04_method.tex
\section{PoE-Fuse}
\label{sec:method}

\begin{figure}[tb]
  \centering
  \includegraphics[width=\linewidth]{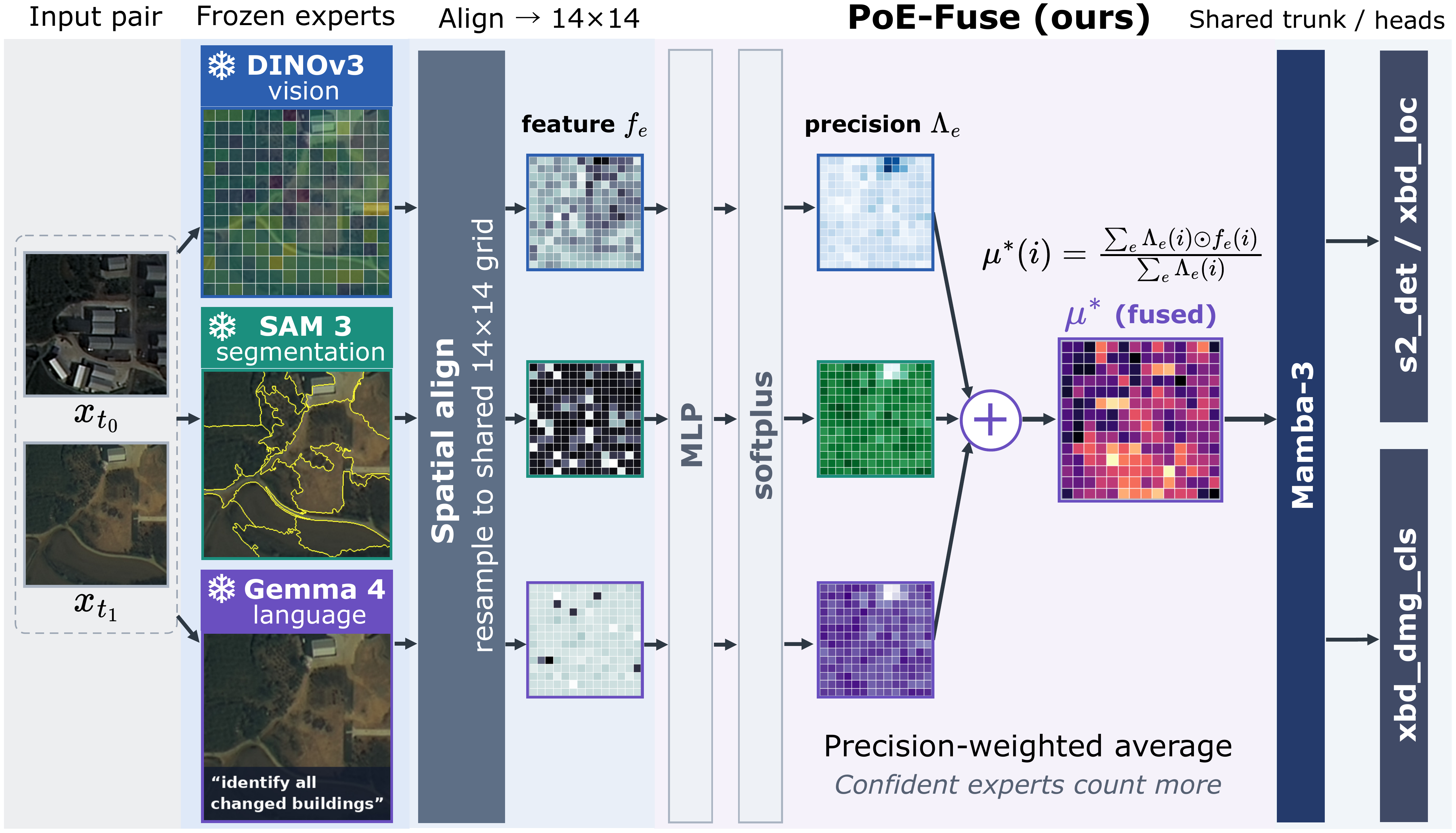}
  \caption{Overview of \method{}.
    Each frozen expert encodes the bi-temporal pair $(x_{t_0},x_{t_1})$ into its own token space.
    Learnable grid queries resample every order-free expert onto a shared $G{\times}G$ grid anchored by the geometric expert; the aligned features and their bi-temporal differences are fused by learned per-cell precision; a lightweight Mamba-3 mixer integrates the fused sequence; and a per-task head reads either the mixed tokens or the dense change field.
    Only the projections, alignment, precision, mixer, and heads are trained; all experts stay frozen.
  }
  \label{fig:method}
\end{figure}

\subsection{Overview and Notation}
\label{sec:method:overview}

\paragraph{Product-of-experts principle.}
\method{} follows the product-of-experts view of fusion~\cite{hinton2002poe}: each frozen expert offers an independent, noisy observation of the same scene, and their combination should trust each expert exactly where it is reliable.
For Gaussian observations this product has a closed form, the precision-weighted average, which we adopt as our fusion rule (Sec.~\ref{sec:method:align}).
Two obstacles separate this principle from practice: the experts' token spaces are mutually incompatible, so ``the same location'' is not even defined across experts, and their reliability varies from cell to cell and from image to image.
We remove the first obstacle by resampling all experts onto the grid of a geometric anchor, and the second by learning per-cell precision end-to-end.

\paragraph{Frozen experts.}
We use $E$ frozen foundation models indexed by $e \in \mathcal{E}$: DINOv3-SAT~\cite{simeoni2025dinov3} for geometric structure, SAM~3~\cite{carion2025sam3} for text-conditioned grounding, and Gemma~4~\cite{gemma2026gemma4} for language semantics; the alignment is backbone-agnostic, and any subset can be selected (Sec.~\ref{sec:exp:ablation}). Their inputs differ: DINOv3-SAT and SAM~3 encode each image, SAM~3 conditioned on a fixed per-task noun phrase (``building'', or ``damaged building'' for damage classification), while Gemma~4 receives the image and the per-sample task prompt through its chat template; we keep the penultimate-layer states of its image-token positions.
For the image acquired at timestamp $t$, expert $e$ emits $N_e$ penultimate tokens of width $C_e$, both of which differ across experts, and a bias-free per-expert linear projection maps them to a shared channel width $d_s$ (all values in Sec.~\ref{sec:exp:details}).
Projection unifies channels but neither token count nor geometry: only the DINOv3-SAT tokens lie on an ordered patch grid, while SAM~3 queries and Gemma~4 soft tokens are order-free (Table~\ref{tab:experts}), so their concatenation has no shared spatial frame.

\begin{table}[tb]
  \centering
  \caption{Frozen experts used by \method{}. Only the geometric encoder carries an explicit spatial layout and serves as the alignment anchor. Token counts and channel widths are given in Sec.~\ref{sec:exp:details}.}
  \label{tab:experts}
  \footnotesize
  \setlength{\tabcolsep}{3pt}
  \begin{tabular}{@{}l p{2.35cm} c p{1.95cm}@{}}
    \toprule
    Expert & Role & Tokens & Layout \\
    \midrule
    DINOv3-SAT~\cite{simeoni2025dinov3} & Geometry / structure & $N_{\mathrm{dino}}$ & Ordered grid (anchor) \\
    SAM~3~\cite{carion2025sam3}         & Text-conditioned grounding & $N_{\mathrm{sam3}}$ & Order-free queries \\
    Gemma~4~\cite{gemma2026gemma4}      & Language semantics & $N_{\mathrm{gemma}}$ & Order-free tokens \\
    \bottomrule
  \end{tabular}
\end{table}

\subsection{Precision-Weighted Expert Fusion on a Shared Grid}
\label{sec:method:align}

\paragraph{Anchored spatial alignment.}
Dense bi-temporal prediction requires the experts to be comparable per location, yet only the geometric expert carries a native spatial layout, so we adopt its $G{\times}G$ patch grid as the common frame rather than inventing a new one.
A set of $G^2$ learnable grid queries, one per cell, resamples each order-free expert onto this grid through pre-norm multi-head cross-attention over the expert's projected tokens, and a learned per-expert tag is added so that downstream layers can distinguish the sources.
The anchor itself bypasses resampling, since its $G^2$ tokens already lie on the grid; this identity anchoring supplies spatial structure from the first training step, and removing it collapses dense prediction (Sec.~\ref{sec:exp:ablation}).
We write $f_e^{t}(i) \in \mathbb{R}^{d_s}$ for the aligned feature of expert $e$ at grid cell $i \in \{1,\dots,G^2\}$ and timestamp $t$.

\paragraph{Gaussian observation model.}
Aligned experts are comparable per cell but not equally reliable (Sec.~\ref{sec:intro}), so we model every aligned feature as a noisy observation of a latent scene state,
\begin{equation}
  f_e^{t}(i) \;=\; \mu^{t}(i) + \varepsilon_e^{t}(i), \qquad \varepsilon_e^{t}(i) \sim \mathcal{N}\!\left(0,\ \operatorname{diag}\!\left(\Lambda_e^{t}(i)\right)^{-1}\right),
  \label{eq:obs}
\end{equation}
where $\mu^{t}(i) \in \mathbb{R}^{d_s}$ is the latent state of cell $i$ at timestamp $t$ and $\Lambda_e^{t}(i) \in \mathbb{R}_{>0}^{d_s}$ is a per-channel precision (inverse variance) that encodes how much expert $e$ is to be trusted at that cell.

\paragraph{Fusion as a product of experts.}
Under Eq.~\eqref{eq:obs} the product of the experts' Gaussian likelihoods is again Gaussian, and its mean, the minimum-variance unbiased estimate of the latent state under this observation model, is the precision-weighted average,
\begin{equation}
  \mu^{*t}(i) \;=\; \Big(\sum\nolimits_{e \in \mathcal{E}} \Lambda_e^{t}(i)\Big)^{-1} \sum\nolimits_{e \in \mathcal{E}} \Lambda_e^{t}(i)\, f_e^{t}(i),
  \label{eq:blue}
\end{equation}
where the sums run over the active experts and all products and inverses are taken elementwise per channel.
This estimator strictly generalizes common fusion rules: a constant $\Lambda_e^{t}(i) \equiv 1$ reduces to the uniform average (naive summation up to a constant scale), and a per-expert constant $\Lambda_e^{t}(i) \equiv g_e$ recovers scalar gating, so both baselines are special cases that our ablation revisits (Sec.~\ref{sec:exp:ablation}).

\paragraph{Learned per-cell precision.}
The true observation noise is unknown, so we learn the precision from the task loss, as in heteroscedastic uncertainty estimation~\cite{kendall2017uncertainty},
\begin{equation}
  \Lambda_e^{t}(i) \;=\; \operatorname{softplus}\!\left(\beta_e + s \cdot \mathrm{MLP}_e\!\left(f_e^{t}(i)\right)\right),
  \label{eq:learned_prec}
\end{equation}
where $\beta_e \in \mathbb{R}$ is a learnable per-expert bias, $s$ is a learnable scale initialized at $0.1$, and $\mathrm{MLP}_e$ is a small per-expert network with $d_s$-dimensional output whose final layer is zero-initialized, so training starts from near-uniform fusion while gradients flow from the first step (MLP size in Sec.~\ref{sec:exp:details}).
We do not claim calibrated uncertainty: $\Lambda$ is shaped only by the task loss, and the assumed independence of errors across experts and timestamps is an idealization, since the experts view the same scene; we therefore treat the estimator as a structured parameterization of fusion weights, validated empirically rather than probabilistically (Sec.~\ref{sec:exp:ablation}).

\paragraph{Difference precision for change fields.}
Change tasks read the bi-temporal difference $d_e(i) = f_e^{t_1}(i) - f_e^{t_0}(i)$, whose per-channel variance under Eq.~\eqref{eq:obs} is the sum $\Lambda_e^{t_0}(i)^{-1} + \Lambda_e^{t_1}(i)^{-1}$ for independent noise.
The precision of the difference is therefore the harmonic composition
\begin{equation}
  \Lambda_e^{\mathrm{diff}}(i) \;=\; \left(\Lambda_e^{t_0}(i)^{-1} + \Lambda_e^{t_1}(i)^{-1}\right)^{-1},
  \label{eq:diff_prec}
\end{equation}
and applying Eq.~\eqref{eq:blue} to $\big(d_e(i), \Lambda_e^{\mathrm{diff}}(i)\big)$ yields the fused change field.
A single low-precision timestamp thus down-weights the whole difference, suppressing spurious change caused by illumination or seasonal variation in either image.

\paragraph{Two coupling sites.}
The trunk consumes the fusion at two sites: the fused change field drives the dense segmentation heads, and the fused per-timestamp features $\mu^{*t}$ drive the mean-pooled classifier head.
Weighting both sites is complementary, and restricting the fusion to either one loses accuracy (Sec.~\ref{sec:exp:ablation}).

\subsection{Lightweight Mamba-3 Mixer}
\label{sec:method:mixer}

The fused per-timestamp grids and change field are flattened in raster order, tagged with learnable temporal embeddings, and mixed by a stack of selective state-space (Mamba-3\rev{~\cite{lahoti2026mamba3}, building on Mamba~\cite{gu2024mamba}}) blocks in pre-norm residual layout, at linear cost in sequence length.
Per-task heads, the only task-dependent part, read the result (Appendix~A).

%% file: tex/sections/05_experiments.tex
\section{Experiments}
\label{sec:experiments}

\subsection{Datasets}
\label{sec:exp:datasets}

We evaluated on the three tasks of Sec.~\ref{sec:problem} using the official TEOChatlas splits~\cite{irvin2025teochat} without new annotations (Table~\ref{tab:dataset}, Fig.~\ref{fig:teaser}).
\rev{Each damage-classification record pairs one queried building with its own crop, which already localizes it; as a control, deranging the query coordinates across records leaves our damage F1 unchanged ($54.8 \pm 3.6$ against $54.8 \pm 3.5$, five seeds).}

\input{tex/tables/tab_dataset_stats}

\subsection{Evaluation Metrics}
\label{sec:exp:metrics}

Following the TEOChat evaluation code, we rasterized polygon annotations to $256{\times}256$ binary masks and computed \emph{pixel-level F1} at a fixed sigmoid threshold of $0.5$, not tuned on the eval split, for change detection and building localization; our $32{\times}32$ head logits are bilinearly upsampled to this resolution before thresholding.
For damage classification we reported the \emph{inverse-prevalence-weighted F1}, averaging per-class F1 (over the five xBD damage labels) with weights inversely proportional to class frequency.

\subsection{Implementation Details}
\label{sec:exp:details}

Inputs are resized to $224{\times}224$; the experts emit $N_{\mathrm{dino}}{=}196$ ($14{\times}14$), $N_{\mathrm{sam3}}{=}200$, and $N_{\mathrm{gemma}}{=}256$ tokens with widths $C_e = 4096$, $256$, and $2560$, projected to the shared width $d_s{=}1024$ on the $G{=}14$ anchor grid; the precision MLP has two layers; the mixer stacks four Mamba-3 blocks with state size $128$ and expansion factor $2$.
The expert features depend only on the input, so we computed them once and cached them; training then optimized only the projections, alignment, fusion, mixer, and heads (59.7M trainable parameters for the shared trunk), about $118\times$ fewer trained parameters than instruction-tuning the 7B vision-language baseline, an efficiency that concerns adaptation, not deployment: at inference every frozen expert runs (15.6B activated parameters), and the measured latency of \rev{276\,ms per pair is comparable to, though slightly above,} the baseline's 114--260\,ms per sample \rev{(per-component latency, FLOPs, memory, and cache size in Appendix~A)}.

A single shared trunk was trained jointly on the three change-detection tasks with PyTorch and AdamW on a single NVIDIA RTX PRO~6000 GPU; the damage-classification head used a focal cross-entropy ($\gamma{=}1.5$) with inverse-frequency class weights to counter the dominant ``no damage'' class.
All main results and fusion-rule ablations are reported as mean\,$\pm$\,std over five seeds (42, 123, 456, 789, 999); the expert-subset and coupling-site ablations use three seeds (42, 123, 456).
\rev{For every trunk run we report the epoch with the best three-task mean on the eval split (patience 8); selecting the last epoch instead changes the dense results little (46.0 and 74.5 pixel-F1 on change detection and localization over four seeds, against 46.9 and 75.6 for the selected epochs), whereas damage classification is unstable across epochs and its value depends on this selection.}
\rev{The grid size $G{=}14$ is the native patch grid of the DINOv3-SAT anchor at $224{\times}224$ input rather than a tuned value; moving to $G{=}28$ removes that anchor and training fails to descend (three-task mean 15.9 at seed 42, the same anchor-removal collapse as in Table~\ref{tab:component}). Four Mamba-3 blocks are used for stability over two ($58.8 \pm 0.8$ against $55.1 \pm 7.6$ mean F1 over three seeds), and a self-attention mixer performs on par (Table~\ref{tab:ablation}). The shared width $d_s{=}1024$ lies between the expert widths of 256 and 4096, the precision head is zero-initialized so that training starts from near-uniform fusion, and the penultimate-layer tokens, the per-task noun phrases for SAM~3, and the per-sample prompts for Gemma~4 follow Sec.~\ref{sec:method} without ablation.}

\input{tex/tables/tab_main_results}

\begin{figure}[!tp]
  \centering
  \includegraphics[width=\linewidth]{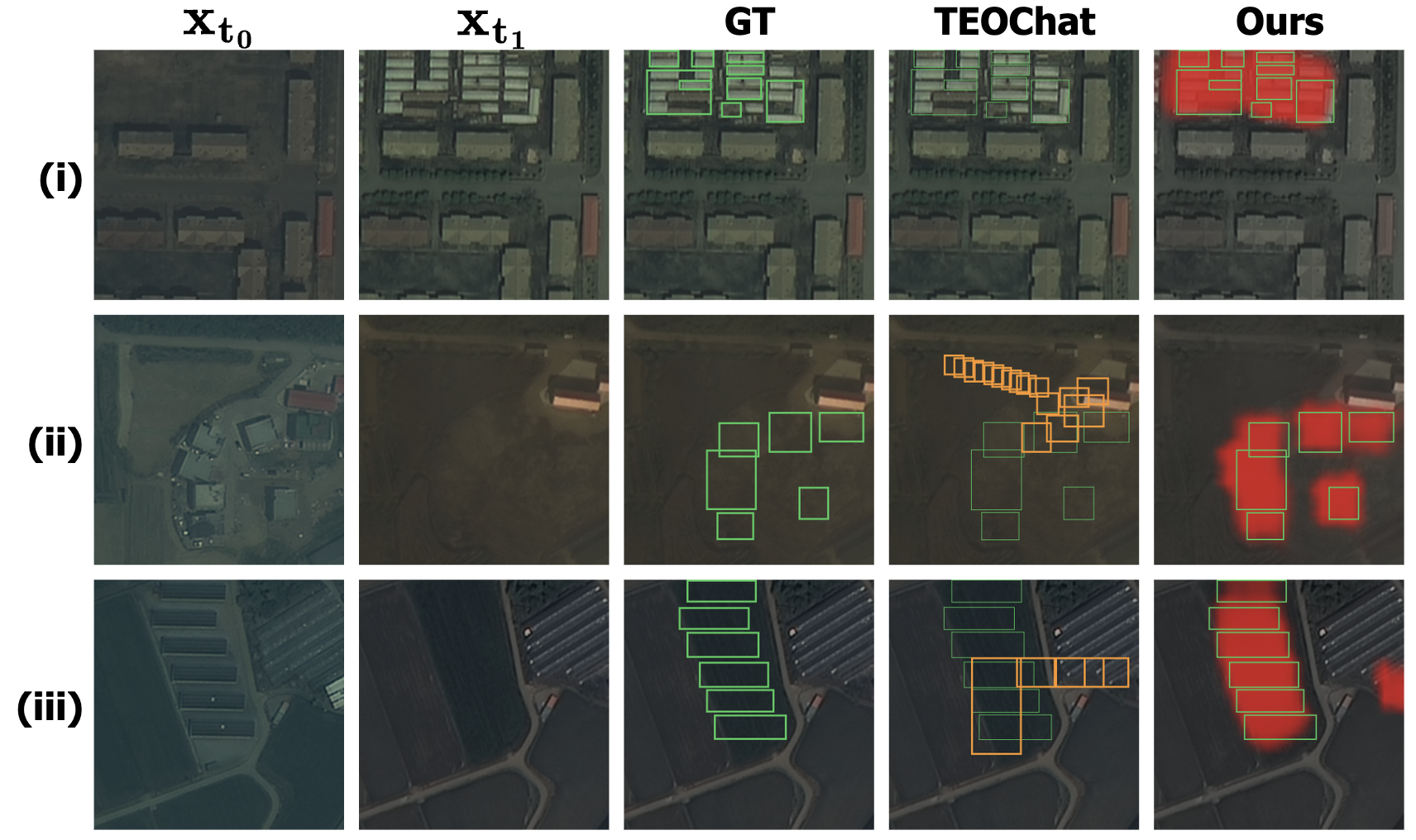}\\[4pt]
  \includegraphics[width=\linewidth]{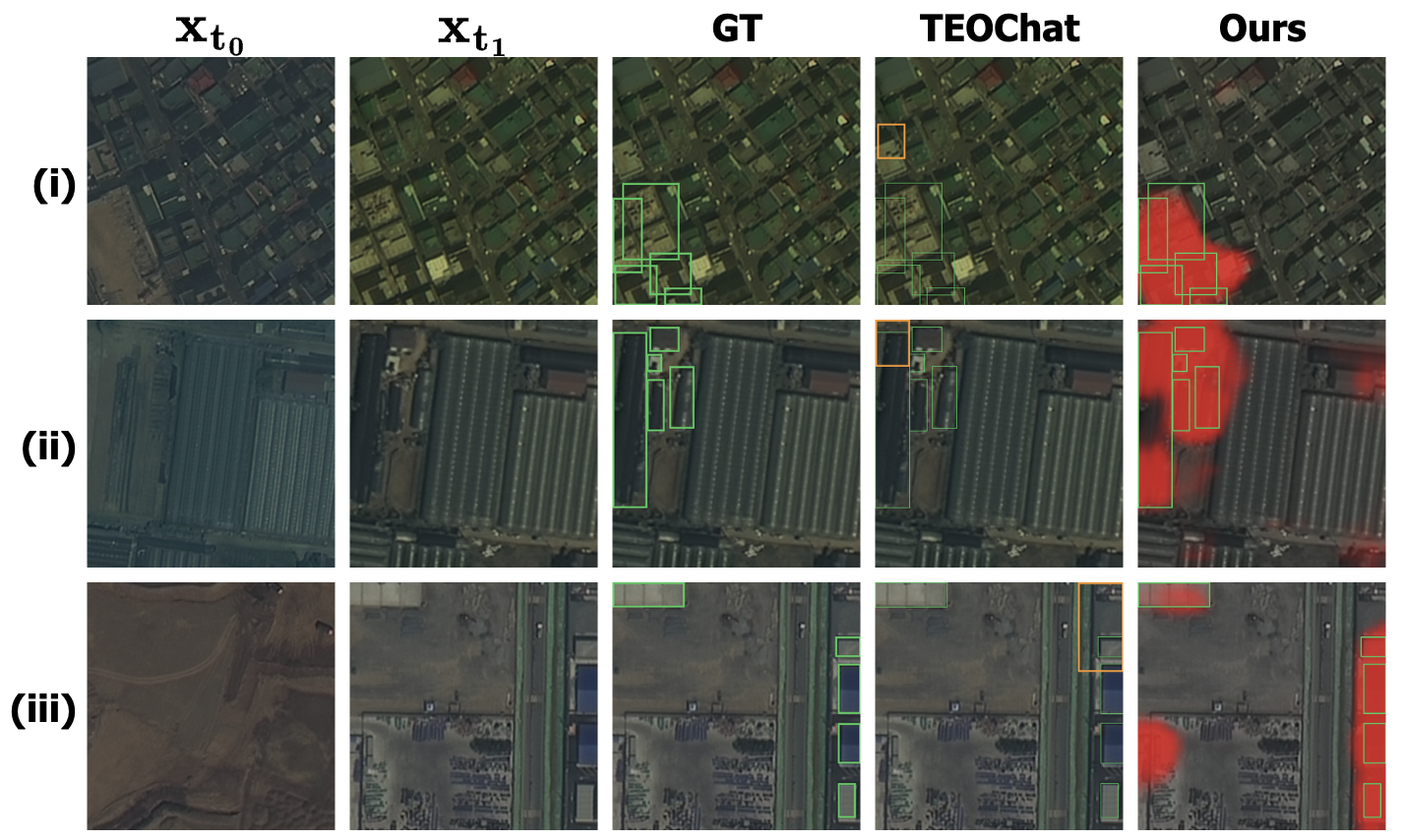}
  \caption{Qualitative comparison on S2Looking change detection.
    Columns: pre-event image $x_{t_0}$, post-event image $x_{t_1}$, ground truth, TEOChat, and \method{} (Ours).
    Green boxes mark ground-truth change, orange boxes are TEOChat predictions, and the red overlay is our predicted change field.
    TEOChat emits sparse, displaced boxes that miss many changed buildings, whereas \method{} produces dense change masks that adhere to the ground-truth footprints.
    Additional examples appear in Appendix~B.}
  \label{fig:qualitative}
\end{figure}

\subsection{Baselines}
\label{sec:exp:baselines}

We compared against two families of prior work, both evaluated with the identical pixel-F1 harness.
\emph{Specialist} change-detection models span classical and modern designs: FC-Siam-Diff~\cite{daudt2018fcsiam}, SNUNet~\cite{fang2022snunet}, TinyCD~\cite{codegoni2023tinycd}, ChangeFormer~\cite{bandara2022changeformer}, and ChangeMamba~\cite{chen2024changemamba}, each trained with standard change-detection losses (Dice + BCE, \rev{30 epochs, the same budget as our trunk}); of these, only the classical baseline extends to damage classification through an adapted classification variant.
These numbers are not comparable to those reported on the specialists' original benchmarks: each model is trained from scratch on the small TEOChatlas splits ($607$--$2{,}008$ pairs) at $224{\times}224$, so Table~\ref{tab:main} measures all systems under one protocol, not the specialists' ceiling at native resolution and full training scale.
The \emph{generalist} baseline is TEOChat~\cite{irvin2025teochat}, an instruction-tuned temporal vision-language model that, like \method{}, solves all tasks with one model.

\subsection{Main Results}
\label{sec:exp:main}

In Table~\ref{tab:main}, \method{} reaches $47.2 \pm 1.3\%$ pixel-F1 on S2Looking change detection, $75.4 \pm 1.6\%$ on xBD localization, and $54.8 \pm 3.5\%$ inverse-prevalence-weighted F1 on damage classification, for a three-task mean of $59.2\%$, $18.5$ points above the generalist VLM ($40.7\%$) while training only ${\sim}$60M parameters on cached features.
It surpasses all five specialist CD models on both dense tasks \rev{under a matched training budget: retrained for the same 30 epochs, the strongest specialist reaches $46.6 \pm 1.2\%$ on change detection and $73.0 \pm 2.0\%$ on localization (both TinyCD, three seeds), below} \method{}.
\rev{The ranking over the generalist baseline is preserved under IoU ($30.9$/$60.6$ vs.\ $20.8$/$23.4$) and a 2-px boundary F1 ($13.8$/$28.0$ vs.\ $11.3$/$15.8$) computed on the same masks (Appendix~C), so the fixed-threshold pixel F1 is not what the advantage rests on.}
ChangeMamba, from the same Mamba state-space family as our mixer, reaches only \rev{$38.6\%$ and $64.0\%$ at the same budget}, consistent with the gain stemming from precision-weighted fusion of frozen experts rather than the choice of mixer.
On damage classification, \method{} leads both the adapted classical baseline ($28.3\%$) and the \rev{released} generalist VLM ($54.8\%$ against $49.7\%$).

\subsection{Qualitative Results}
\label{sec:exp:qualitative_fig}

In Fig.~\ref{fig:qualitative}, per-crop pixel-F1 is computed by the same harness, which rasterizes the TEOChat boxes and thresholds our field.
On change detection and localization, TEOChat emits sparse, box-like predictions that miss building boundaries and frequently misplace the change region, yielding high precision but low recall; \method{} instead recovers most ground-truth buildings as dense masks, at the granularity of the $14{\times}14$ grid; its typical failures are over-predictions beyond small ground-truth regions, costing precision rather than recall (Appendix~B).
Per crop, \method{} ranges from $0.54$ to $0.82$ on change detection and $0.71$ to $0.82$ on localization, against $0.0$ to $0.42$ and $0.0$ to $0.25$ for TEOChat on the same crops.
On damage classification (Appendix~B), TEOChat tends to default to the majority ``no damage'' class (about $64\%$ of buildings) and misses the rare destroyed and major-damage cases that matter most for disaster response, whereas \method{} recovers them, lifting its inverse-prevalence-weighted F1.

\input{tex/tables/tab_ablation}
\input{tex/tables/tab_component_ablation}

\subsection{Ablation Study}
\label{sec:exp:ablation}

Table~\ref{tab:ablation} varies only the fusion rule. Its special cases, uniform summation ($\Lambda_e \equiv 1$, $57.1\%$ mean F1) and scalar cross-expert gating ($\Lambda_e \equiv g_e$, $58.1\%$), both fall below \method{} ($59.2\%$)\rev{, although the margin over scalar gating lies within seed noise (seed-paired $p{=}0.37$)}.
Two design choices account for the gap.
First, precision \rev{should} be \emph{dynamic}: a static, input-independent variant retains little of the benefit ($57.4\%$, barely above uniform summation\rev{; $p{=}0.015$}), \rev{which indicates} that the gain comes from adapting trust per location rather than from the extra parameters.
Second, the difference-precision corollary (Eq.~\ref{eq:diff_prec}) contributes: fusing single-timestamp precisions instead drops the mean to $58.2\%$.
The gain is also robust to the mixer: a self-attention mixer reaches $59.0\%$, close to the full model, indicating that precision-weighted fusion rather than the sequence model is the operative component.

Table~\ref{tab:component} further ablates which experts are fused and where the fusion is applied.
Removing the geometric anchor is catastrophic: without DINOv3-SAT, the only expert whose tokens natively form the shared grid, the trunk fails to learn dense change at all (pixel-F1 $0.0$ on S2Looking, mean $16.0$), confirming that anchored alignment, not merely the number of experts, is what makes fusion effective.
The remaining pairs are complementary: DINO+SAM3 is the strongest pair (slightly ahead on change detection), dropping SAM3 costs $16$ points on damage classification, and \rev{adding Gemma leaves the mean essentially unchanged on the same three seeds ($58.9$ versus $58.8$; $59.2$ over five seeds)}.
Applying precision-weighted fusion at both coupling sites is likewise necessary: restricting it to the dense change field or to the pooled classifier features lowers the mean to $58.5$ and $57.5$ respectively.

%% file: tex/tables/tab_dataset_stats.tex
\begin{table}[tb]
  \caption{Dataset statistics for the three benchmark tasks (TEOChatlas official splits).
    Train counts filter \texttt{train/instruct.json} by dataset and task;
    eval counts use the official eval JSON per task.
    \rev{Each damage-classification eval record pairs one queried building with its own image crop, so the 6{,}322 records are 6{,}322 distinct crops.
    The xBD train/eval asymmetry is inherited from the release: the official instruction-tuning JSON spreads its 19{,}749 xBD examples over many templates and contains only 627/607 instances of these two, whereas the eval JSONs cover the full eval split.}}
  \label{tab:dataset}
  \centering
  \small
  \setlength{\tabcolsep}{4pt}
  \begin{tabular}{@{}lrrl@{}}
    \toprule
    Task & Train & Eval & Notes \\
    \midrule
    S2Looking change detection
      & 2{,}008 & 4{,}798 & per-pixel mask \\
    xBD building localization
      & 627 & 6{,}322 & per-pixel mask \\
    xBD damage classification
      & 607 & 6{,}322 & 5-way label \\
    \bottomrule
  \end{tabular}
\end{table}

%% file: tex/tables/tab_main_results.tex
\begin{table}[tb]
  \caption{Main results (same pixel-F1 harness for all systems; Sec.~\ref{sec:exp:baselines}): mean\,$\pm$\,std over five seeds for \method{} \rev{and TEOChat, and over three seeds (0, 1, 2) for the specialists, which are retrained for 30 epochs to match the budget of our trunk}.
    \emph{Specialist}: dedicated change-detection models, each trained \emph{separately per task} on the dense tasks (the classical baseline also on damage classification via an adapted variant).
    \emph{Generalist}: a \emph{single} model handles all three tasks.
    S2Look.\ det.\ and xBD loc.\ report pixel-F1, xBD dmg.\ inverse-prevalence-weighted F1; \emph{Mean} is reported only for single models spanning all tasks.
    All per-task gains of \method{} over TEOChat are statistically significant under Welch's $t$-test across seeds ($p<10^{-4}$ on the dense tasks, $p<0.05$ on damage classification).}
  \label{tab:main}
  \centering
  \small
  \setlength{\tabcolsep}{5pt}
  \begin{tabular}{@{}lcccc@{}}
    \toprule
    System & Mean\,$\uparrow$ & S2Look.\ det.\,$\uparrow$ & xBD loc.\,$\uparrow$ & xBD dmg.\,$\uparrow$ \\
    \midrule
    \rowcolor{black!8}[0pt][0pt]
    \multicolumn{5}{@{}l@{}}{\textit{Specialist}} \\
    ChangeFormer~\cite{bandara2022changeformer} & --- & \rev{37.7 $\pm$ 0.6} & \rev{52.9 $\pm$ 1.9} & --- \\
    ChangeMamba~\cite{chen2024changemamba}      & --- & \rev{38.6 $\pm$ 2.4} & \rev{64.0 $\pm$ 1.3} & --- \\
    TinyCD~\cite{codegoni2023tinycd}            & --- & \rev{46.6 $\pm$ 1.2} & \rev{73.0 $\pm$ 2.0} & --- \\
    FC-Siam-Diff~\cite{daudt2018fcsiam}         & --- & \rev{45.1 $\pm$ 1.7} & \rev{72.2 $\pm$ 0.7} & 28.3 $\pm$ 1.2 \\
    SNUNet~\cite{fang2022snunet}                & --- & \rev{44.1 $\pm$ 2.1} & \rev{72.3 $\pm$ 1.2} & --- \\
    \midrule
    \rowcolor{black!8}[0pt][0pt]
    \multicolumn{5}{@{}l@{}}{\textit{Generalist}} \\
    TEOChat~\cite{irvin2025teochat}             & 40.7 & 34.4 $\pm$ 0.4 & 38.0 $\pm$ 0.2 & 49.7 $\pm$ 0.2 \\
    \midrule
    \rev{Ours (full system)} & \textbf{59.2} & \textbf{47.2 $\pm$ 1.3} & \textbf{75.4 $\pm$ 1.6} & \textbf{54.8 $\pm$ 3.5} \\
    \bottomrule
  \end{tabular}
\end{table}

%% file: tex/tables/tab_ablation.tex
\begin{table}[tb]
  \caption{Ablation on the fusion rule: the shared trunk is fixed and only the expert-combination rule changes.
    \rev{Per-task and three-task mean F1} ($\times100$), mean\,$\pm$\,std over \rev{the same} five seeds \rev{for every row}.
    \rev{Seed-paired two-sided $t$-tests on the mean against \method{}: uniform sum $p{=}0.047$, static precision $p{=}0.015$ (5 of 5 seeds), scalar gate $p{=}0.37$, no corollary $p{=}0.24$, transformer mixer $p{=}0.82$, softmax gate $p{=}0.23$.
    Second block: alternative fusion rules on the same trunk (fusion-rule parameters in parentheses, \method{} 6.30M); seed-and-sample bootstrap $p{=}0.002$ (concatenation) and $0.004$ (cross-attention).}}
  \label{tab:ablation}
  \centering
  \scriptsize
  \setlength{\tabcolsep}{3pt}
  \begin{tabular}{@{}llcccc@{}}
    \toprule
    Fusion rule & Precision $\Lambda_e$ & \rev{S2Look.\ det.} & \rev{xBD loc.} & \rev{xBD dmg.} & Mean \\
    \midrule
    uniform sum            & $\Lambda_e \equiv 1$         & \rev{46.0 $\pm$ 0.6} & \rev{74.4 $\pm$ 1.3} & \rev{50.8 $\pm$ 1.9} & \rev{57.1 $\pm$ 1.1} \\
    scalar gate            & $\Lambda_e \equiv g_e$       & \rev{47.2 $\pm$ 2.9} & \rev{73.8 $\pm$ 2.1} & \rev{53.3 $\pm$ 1.6} & \rev{58.1 $\pm$ 1.1} \\
    static precision       & input-independent           & \rev{46.9 $\pm$ 1.8} & \rev{73.9 $\pm$ 0.5} & \rev{51.3 $\pm$ 2.3} & \rev{57.4 $\pm$ 0.8} \\
    no corollary           & PoE, single timestamp       & \rev{46.6 $\pm$ 2.2} & \rev{74.5 $\pm$ 2.2} & \rev{53.5 $\pm$ 2.3} & \rev{58.2 $\pm$ 1.5} \\
    transformer mixer      & PoE + self-attention        & \rev{48.8 $\pm$ 2.9} & \rev{76.5 $\pm$ 2.1} & \rev{51.6 $\pm$ 4.1} & \rev{59.0 $\pm$ 1.0} \\
    \midrule
    \rev{concatenation} & \rev{concat+linear (6.29M)}     & \rev{45.8 $\pm$ 3.7} & \rev{70.4 $\pm$ 2.6} & \rev{49.2 $\pm$ 2.7} & \rev{55.1 $\pm$ 2.5} \\
    \rev{cross-attention} & \rev{8.40M}                     & \rev{46.1 $\pm$ 1.7} & \rev{70.5 $\pm$ 1.8} & \rev{40.2 $\pm$ 13.5} & \rev{52.3 $\pm$ 5.2} \\
    \rev{softmax gate} & \rev{per cell/channel (6.30M)} & \rev{46.1 $\pm$ 3.2} & \rev{73.5 $\pm$ 2.3} & \rev{52.5 $\pm$ 2.0} & \rev{57.4 $\pm$ 1.5} \\
    \midrule
    \method{} (Ours)       & full per-cell dynamic        & \rev{47.2 $\pm$ 1.3} & \rev{75.4 $\pm$ 1.6} & \rev{\textbf{54.8 $\pm$ 3.5}} & \textbf{59.2 $\pm$ \rev{1.2}} \\
    \bottomrule
  \end{tabular}
\end{table}

%% file: tex/tables/tab_component_ablation.tex
\begin{table}[tb]
  \caption{Component ablations on the shared trunk: mean over three seeds (42/123/456); the full model (last row) uses five.
    Left: which frozen experts are fused; \emph{PoE site} is where precision-weighted fusion is applied, with the other coupling site reverting to gated summation.
    \rev{$^{\dagger}$Single-expert rows; the same three seeds give 58.8 for the full model.}
    Best per column in bold.}
  \label{tab:component}
  \centering
  \footnotesize
  \setlength{\tabcolsep}{3.5pt}
  \begin{tabular}{@{}ccclcccc@{}}
    \toprule
    \multicolumn{3}{c}{Experts} & & \multicolumn{4}{c}{F1 ($\times$100)} \\
    \cmidrule(r){1-3} \cmidrule(l){5-8}
    DINO & SAM~3 & Gemma & PoE site & Mean\,$\uparrow$ & S2Look.\,$\uparrow$ & xBD loc.\,$\uparrow$ & xBD dmg.\,$\uparrow$ \\
    \midrule
    \rev{\checkmark} &            &            & \rev{Both} & \rev{58.1$^{\dagger}$} & \rev{48.7} & \rev{73.7} & \rev{51.8} \\
               & \rev{\checkmark} &            & \rev{Both} & \rev{24.7$^{\dagger}$} & \rev{9.5} & \rev{43.8} & \rev{21.0} \\
               &            & \rev{\checkmark} & \rev{Both} & \rev{18.6$^{\dagger}$} & \rev{2.7} & \rev{36.0} & \rev{17.2} \\
    \midrule
    \checkmark & \checkmark &            & Both & 58.9 & \textbf{48.9} & 74.6 & 53.0 \\
    \checkmark &            & \checkmark & Both & 52.9 & 47.8 & 72.1 & 38.7 \\
               & \checkmark & \checkmark & Both & 16.0 & 0.0  & 34.6 & 13.3 \\
    \midrule
    \checkmark & \checkmark & \checkmark & Dense      & 58.5 & 46.8 & 74.8 & 53.8 \\
    \checkmark & \checkmark & \checkmark & Classifier & 57.5 & 47.0 & 74.1 & 51.4 \\
    \midrule
    \checkmark & \checkmark & \checkmark & Both & \textbf{59.2} & 47.2 & \textbf{75.4} & \textbf{54.8} \\
    \bottomrule
  \end{tabular}
\end{table}

%% file: tex/sections/06_conclusion.tex
\section{Conclusion}
\label{sec:conclusion}

We studied bi-temporal change understanding on three tasks of the TEOChat benchmark (change detection, building localization, and damage classification) and proposed \method{}, which treats the aligned features of heterogeneous frozen foundation experts as Gaussian observations of a shared latent state and fuses them by their learned per-cell precision, the best linear unbiased estimator.

Across five seeds, a single shared trunk built on this fusion, training only ${\sim}$60M parameters on cached penultimate features, reached $47.2\%$ on S2Looking change detection, $75.4\%$ on xBD localization, and $54.8\%$ on damage classification (mean $59.2\%$), surpassing all five dedicated change-detection baselines we evaluated \rev{when retrained under the same protocol and training budget} and an instruction-tuned temporal vision-language assistant (mean $40.7\%$).
Ablations traced these results to three design choices: the fusion strictly generalizes uniform summation and scalar gating and improves over \rev{the former, with the margin over the latter within seed noise, and} the difference-precision corollary \rev{adds} a further margin on change fields; the geometric anchor is indispensable, as removing it collapses change detection to zero pixel-F1, \rev{the grounding expert is essential for damage classification, and the language expert adds little on the same seeds}; and precision fusion is needed at both coupling sites, the dense change field and the pooled classifier features.

Our study has limitations: inputs are fixed to co-registered pairs at a single resolution, the $14{\times}14$ alignment grid bounds the spatial granularity of the predicted masks, the dense specialist baselines are trained from scratch on small splits and do not extend to categorical damage classification, and although training is light, inference still runs every frozen expert, so deployment cost is dominated by the largest backbone.
Future work includes higher-resolution alignment, calibrated rather than task-driven precision, longer or unregistered acquisition sequences, and extending the trunk to language-queried referring segmentation.

%% file: tex/sections/supp_body.tex
\appendix
\suppressfloats[t]

\section{Task-Specific Heads and Training}
\label{sec:appendix:heads}

\subsection{Prediction Heads}

The mixed token sequence from the Mamba-3 stack feeds into task-specific prediction heads; the dense change field is available to segmentation heads.
We train one head per task while sharing the fusion trunk (projections, alignment, precision fusion, and mixer).

\paragraph{Damage classification.}
A classifier head mean-pools valid tokens of the mixed sequence, applies layer normalization, and predicts five-way damage logits via a linear layer.
The queried building is specified in the task prompt as normalized box coordinates, following the TEOChatlas protocol, and reaches the model through the language expert; the visual input remains the full $224{\times}224$ pair.
We use a focal cross-entropy ($\gamma{=}1.5$) with inverse-frequency class weights to mitigate majority-class collapse.

\paragraph{Dense change segmentation.}
A convolutional head reads the $(G,G,d_s)$ change field and applies a $1{\times}1$ convolution, a bilinear resize to $(32,32)$, a small $3{\times}3$ convolutional body with group normalization and GELU, and a $1{\times}1$ output layer to produce per-pixel logits.
At evaluation the $32{\times}32$ logits are bilinearly upsampled to the $256{\times}256$ mask resolution used by the benchmark scorer.
Loss combines positive-weighted binary cross-entropy and soft Dice regularization.

\subsection{Training Setup}

Training optimizes cached frozen-expert features only; gradients flow into the per-expert projections, alignment module, precision heads, Mamba-3 stack, and task head.
We use AdamW with learning rate $5 \times 10^{-4}$, weight decay $10^{-4}$, and a single NVIDIA RTX PRO 6000 Blackwell GPU (96 GB).
Early stopping (patience 8) on validation pixel-F1 for dense tasks; damage classification runs for exactly 5 epochs with the focal cross-entropy above to stabilize imbalanced predictions.
Batch sizes: 8 (dense), 2 (damage classification).
Typical wall-clock time per task: 2-4 hours after feature caching.
\rev{Per image pair at batch size 1 and $224{\times}224$ (median of 15 runs on a dedicated GPU), the frozen experts take 270\,ms (DINOv3-SAT 40, SAM~3 114, Gemma~4 116) and the trunk 6\,ms, for 276\,ms in total. The FLOPs per pair are 5451\,G for DINOv3-SAT, 8610\,G for SAM~3, and 55\,G for the trunk (the counter does not cover Gemma~4). Peak GPU memory is 31.4\,GB at inference and 3.4\,GB in training, since training reads cached tensors and runs no expert forward, and the one-off feature cache occupies 117\,GB for the three tasks.}

\begin{figure}[p]
  \centering
  \includegraphics[width=\linewidth]{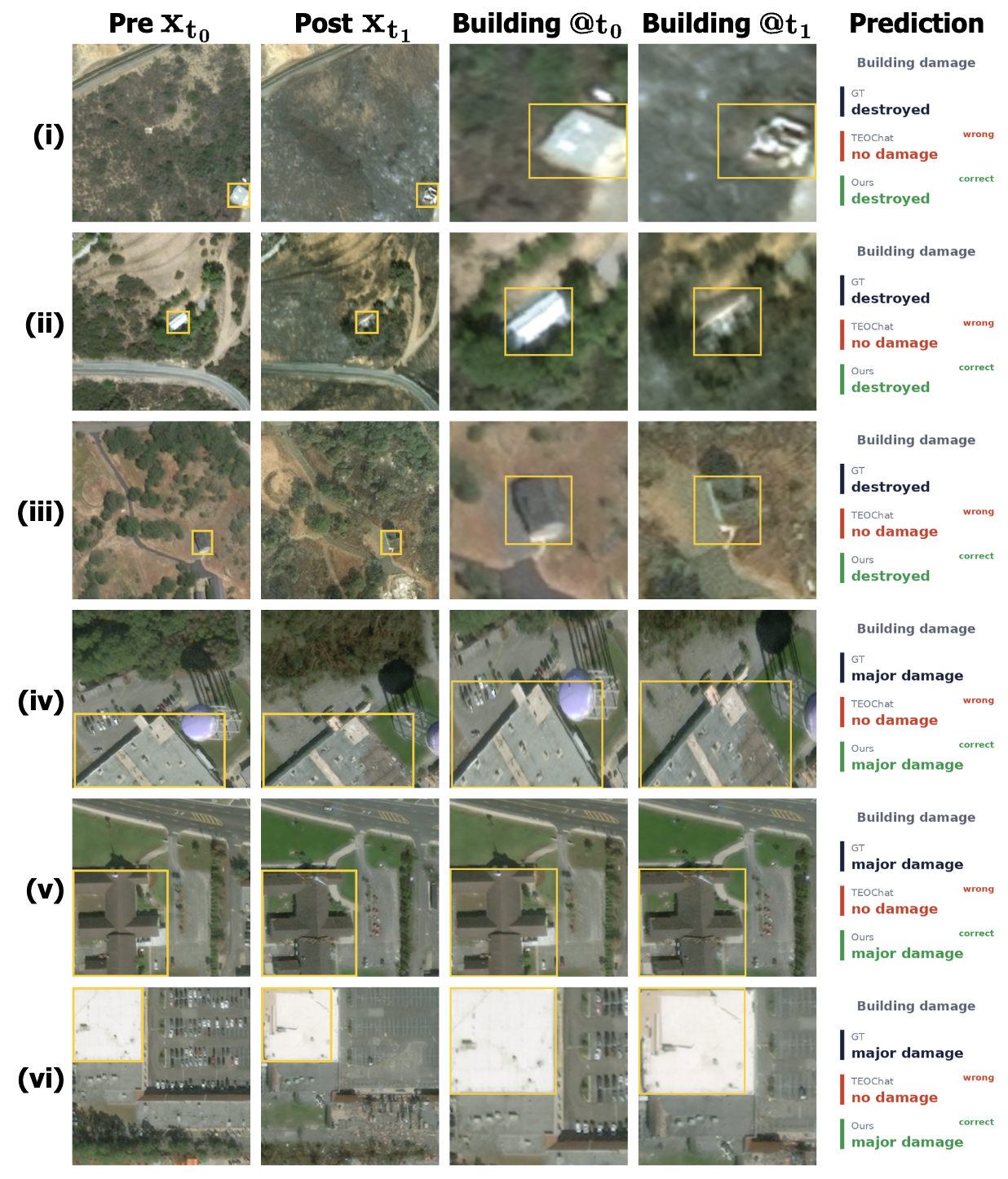}
  \caption{Qualitative building damage classification on xBD.
    Columns: pre-event image $x_{t_0}$, post-event image $x_{t_1}$, the queried building cropped at $t_0$ and $t_1$, and the predicted damage class.
    On these destroyed and major-damage buildings, TEOChat defaults to ``no damage'' (wrong), whereas \method{} predicts the correct severity, which is what drives its higher inverse-prevalence-weighted F1.}
  \label{fig:qual_dmg}
\end{figure}

\begin{figure}[p]
  \centering
  \includegraphics[height=0.41\textheight]{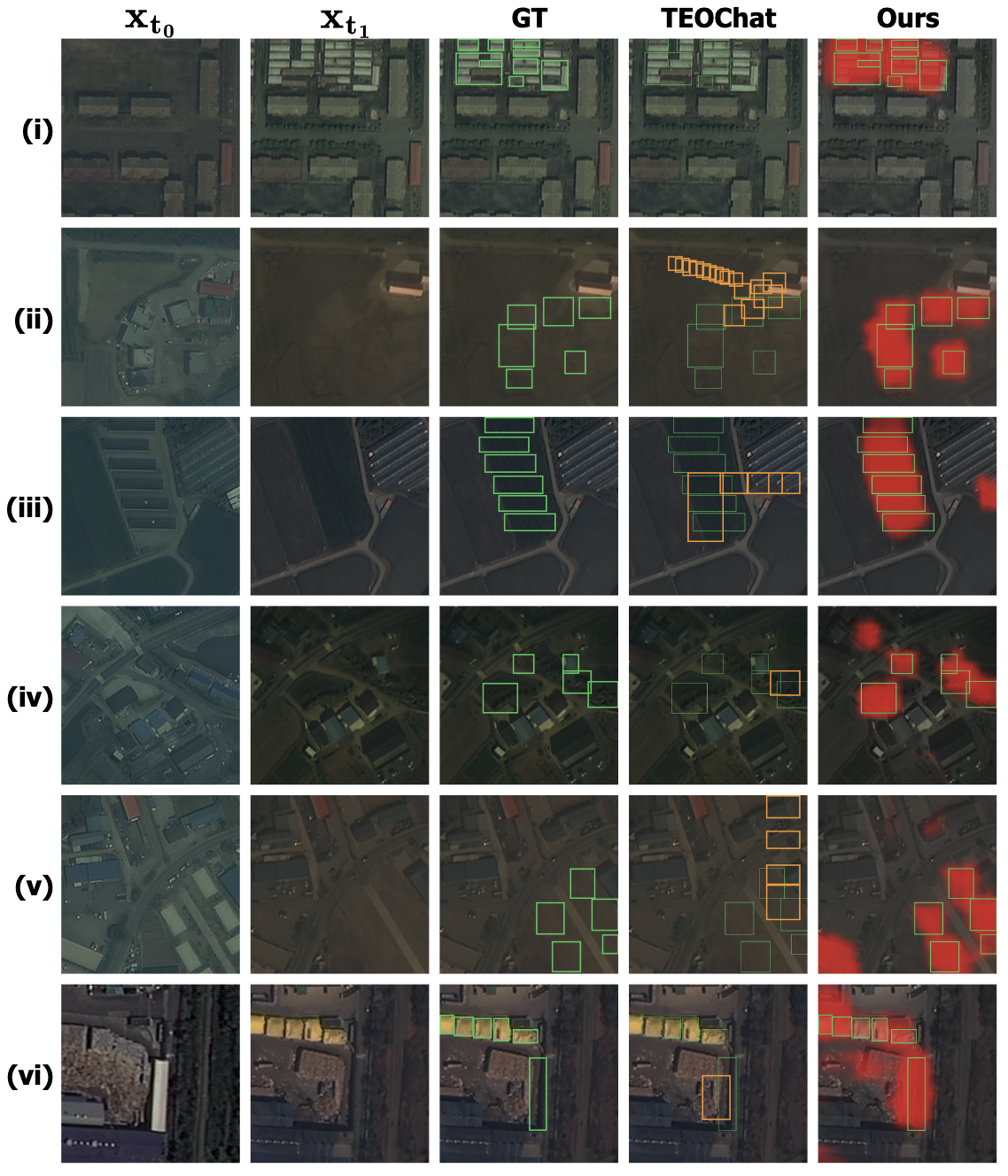}
  \caption{Extended S2Looking change-detection examples (set~1).
    Each row is an independent eval crop with columns $x_{t_0}$, $x_{t_1}$, ground truth, TEOChat, and \method{} (Ours).
    \method{} consistently recovers the changed buildings as a dense change field, while TEOChat's boxes are sparse and frequently displaced.}
  \label{fig:qual_s2_full}
\end{figure}

\begin{figure}[p]
  \centering
  \includegraphics[height=0.41\textheight]{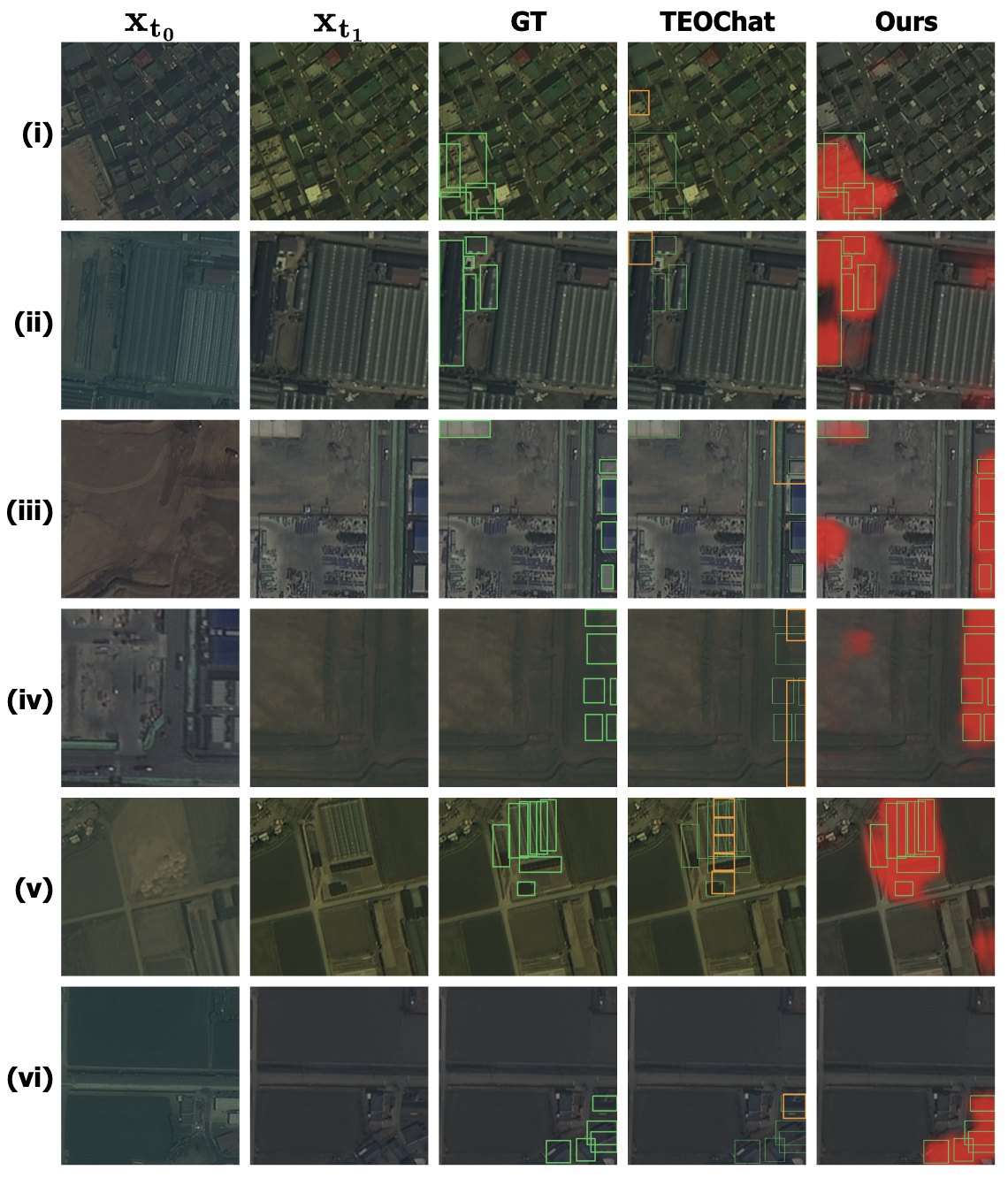}
  \caption{Extended S2Looking change-detection examples (set~2), in the same format as Figure~\ref{fig:qual_s2_full}.}
  \label{fig:qual_s2_p2_full}
\end{figure}
\clearpage

\section{Additional Qualitative Results}
\label{sec:appendix:qual}

We provide extended qualitative comparisons that complement the main-paper examples (Fig.~3).
Figure~\ref{fig:qual_dmg} shows building damage classification, and Figures~\ref{fig:qual_s2_full} and~\ref{fig:qual_s2_p2_full} show additional S2Looking change-detection crops.
The color scheme follows the main paper: green marks ground-truth change, orange marks TEOChat predictions, and the red overlay is our predicted change field.

\rev{%
\section{IoU and Boundary Metrics on the Dense Tasks}
\label{sec:appendix:iou}

Table~\ref{tab:iou_boundary} complements the pixel-F1 comparison of the main paper with IoU and a boundary metric, computed by the same harness on the identical $256{\times}256$ masks: IoU from the official confusion-matrix evaluator, and boundary F1 as the BF score with a 2-px tolerance (boundaries as the XOR of each mask with its 1-px erosion), aggregated at the dataset level.
The ranking under pixel F1 is preserved under both metrics.
The boundary-F1 margin is smallest on S2Looking change detection, consistent with the spatial granularity bound of the $14{\times}14$ alignment grid discussed in the limitations.
}

\begin{table}[!htb]
  \caption{\rev{IoU and boundary F1 ($\times100$, mean\,$\pm$\,std over five seeds) on the two dense tasks, same evaluation harness and masks as the pixel F1 of Table~3. TEOChat xBD localization IoU and boundary F1 are over four seeds.}}
  \label{tab:iou_boundary}
  \centering
  \footnotesize
  \setlength{\tabcolsep}{4pt}
  \begin{tabular}{@{}llccc@{}}
    \toprule
    \rev{System} & \rev{Task} & \rev{Pixel F1} & \rev{IoU} & \rev{Boundary F1} \\
    \midrule
    \rev{TEOChat}        & \rev{S2Look.\ det.} & \rev{34.4 $\pm$ 0.4} & \rev{20.8 $\pm$ 0.3} & \rev{11.3 $\pm$ 0.1} \\
    \rev{TEOChat}        & \rev{xBD loc.}      & \rev{38.0 $\pm$ 0.2} & \rev{23.4 $\pm$ 0.1} & \rev{15.8 $\pm$ 0.1} \\
    \midrule
    \rev{\method{} (Ours)} & \rev{S2Look.\ det.} & \rev{\textbf{47.2 $\pm$ 1.3}} & \rev{\textbf{30.9 $\pm$ 1.1}} & \rev{\textbf{13.8 $\pm$ 0.5}} \\
    \rev{\method{} (Ours)} & \rev{xBD loc.}      & \rev{\textbf{75.4 $\pm$ 1.6}} & \rev{\textbf{60.6 $\pm$ 2.1}} & \rev{\textbf{28.0 $\pm$ 1.9}} \\
    \bottomrule
  \end{tabular}
\end{table}